\documentclass[final]{cvpr}

\usepackage{times}
\usepackage{epsfig}
\usepackage{graphicx}
\usepackage{amsmath}
\usepackage{amssymb}
\usepackage{helvet}   
\usepackage{courier}  
\usepackage{url}      
\usepackage{xcolor}
\usepackage{pifont}
\usepackage{booktabs}
\usepackage{multirow}
\usepackage{algorithmic}
\usepackage[linesnumbered,ruled,vlined]{algorithm2e}
\usepackage[export]{adjustbox}
\usepackage{color}
\usepackage{tabularx}
\usepackage[super]{nth}
\usepackage{caption}

\DeclareMathSymbol{\mh}{\mathord}{operators}{`\-}

\usepackage[pagebackref=true,breaklinks=true,colorlinks,bookmarks=false]{hyperref}

\begin{document}

\title{DetMatch: Two Teachers are Better Than One for Joint 2D and 3D Semi-Supervised Object Detection}

\author{Jinhyung Park${}^1{}^\dagger$,~ Chenfeng Xu${}^2$,~ Yiyang Zhou${}^2$,~ Masayoshi Tomizuka${}^2$,~ Wei Zhan${}^2$ \\
${}^1$Carnegie Mellon University ${}^2$ University of California, Berkeley\\
{\tt\small jinhyun1@andrew.cmu.edu~~\{xuchenfeng, yiyang.zhou, tomizuka, wzhan\}@berkeley.edu}
}

\twocolumn[{%
\renewcommand\twocolumn[1][]{#1}%
\maketitle
\begin{center}
\centering
\captionsetup{type=figure}
\vspace{-2em}
\resizebox{0.97\textwidth}{!}{\includegraphics[width=\linewidth]{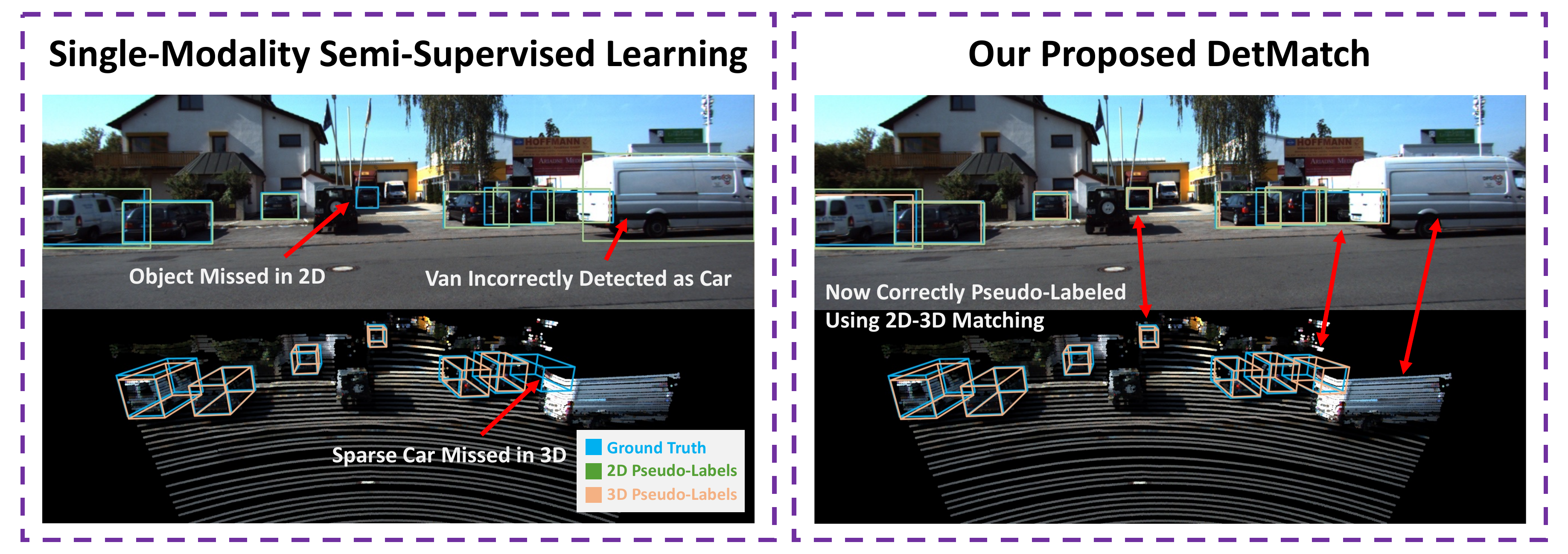}}\vspace{-0.5em}
\captionof{figure}{By matching 2D and 3D detections, our approach resolves false negatives and removes false positives to generate a cleaner set of pseudo-labels. The point cloud is colored for visualization}
\label{fig:teaser}
\vspace{-0.5em}
\end{center}
}]

\setlength{\abovedisplayskip}{0.5\abovedisplayskip}
\setlength{\belowdisplayskip}{0.5\belowdisplayskip}
\setlength{\abovecaptionskip}{0.5\abovecaptionskip}
\setlength{\belowcaptionskip}{0.5\belowcaptionskip}

{\let\thefootnote\relax\footnote{{$^\dagger$Work conducted during visit to University of California, Berkeley.}}}


\begin{abstract}
\vspace{-1em}
While numerous 3D detection works leverage the complementary relationship between RGB images and point clouds, developments in the broader framework of semi-supervised object recognition remain uninfluenced by multi-modal fusion. Current methods develop independent pipelines for 2D and 3D semi-supervised learning despite the availability of paired image and point cloud frames. Observing that the distinct characteristics of each sensor cause them to be biased towards detecting different objects, we propose DetMatch, a flexible framework for joint semi-supervised learning on 2D and 3D modalities. By identifying objects detected in both sensors, our pipeline generates a cleaner, more robust set of pseudo-labels that both demonstrates stronger performance and stymies single-modality error propagation. Further, we leverage the richer semantics of RGB images to rectify incorrect 3D class predictions and improve localization of 3D boxes. Evaluating on the challenging KITTI and Waymo datasets, we improve upon strong semi-supervised learning methods and observe higher quality pseudo-labels. Code will be released \href{https://github.com/Divadi/DetMatch}{here.}
\end{abstract}

\section{Introduction}

Recent advances in Semi-Supervised Learning (SSL) for object recognition focus on the single-modality setting, demonstrating improvements in either 2D or 3D detection when leveraging unlabeled samples of that modality. However, SSL works rarely study the combination of 2D and 3D sensors. In recently published datasets, autonomous vehicles are equipped with a comprehensive collection of sensors that yields multi-modal observations of each scene. Among these devices, 2D RGB cameras and 3D LiDARs have emerged as two independently useful but also mutually complementary modalities. Thus, it is important for SSL methods to utilize both 2D and 3D modalities for autonomous driving applications.

In this work, we propose a novel multi-modal SSL framework, \textbf{DetMatch}, that leverages paired but unlabeled data of multiple modalities to train stronger single-modality object detection models. Our pipeline is agnostic to the specific designs of the 2D and 3D detectors, allowing for flexible usage in conjunction with perpendicular advancements in architectures. 
Further, by yielding single-modality models, our framework does not constrain the trained detectors to the multi-modal or even the autonomous driving setting. 

We observe that differences in modality characteristics between RGB images and point clouds cause them to each be better at detecting different types of objects as illustrated in Figure \ref{fig:teaser}.
3D point clouds are inherently sparse, and the lack of color information causes structurally similar objects to be indistinguishable in the point cloud. On the other hand, 2D RGB images contain a far denser array of color information, allowing for easier discrimination of similarly shaped classes and better detection of objects with few 3D points captured on them. However, unlike point clouds, RGB images lack depth values. Each point in the 3D point cloud represents an exact, observed location in 3D space, making objects spatially separable - this facilitates 3D detection of objects that have overlapping, similar-colored projections in 2D. These factors support our intuition that not only are RGB images and point clouds mutually beneficial, but that their detection results are strongly complementary.

To leverage this relationship for SSL while keeping each detection model single-modal, we associate 2D and 3D results at the detection level. Since 2D and 3D have their own strengths, we use predictions in each modality that have a corresponding detection in the other modality to generate a cleaner subset of box predictions that is used to pseudo-label the unlabeled data for that modality. We find that such pseudo-labels chosen using multiple modalities outperform single-modality generated pseudo-labels. Although this method exploits the advantages of each modality to generate stronger pseudo-labels, it insufficiently utilizes the RGB images' unique rich semantics. In the previous pipeline, a correctly localized \& classified 2D detection cannot directly rectify a poor 3D detection. To remedy this gap, we additionally enforce box and class consistency between matched 2D pseudo-labels and 3D predictions and observe improved performance.

Our main contributions are as follows:
\begin{itemize}
    \item We observe that differences in characteristics between 2D and 3D modalities allow objects of high occlusion to be better detected in 3D, and objects of similar shape but different class to be better identified and localized in 2D.
    \item Our SSL framework leverages the mutually beneficial relationship between multiple modalities during training to yield stronger single-modality models.
    \item We extensively validate DetMatch the difficult KITTI \cite{Geiger2012AreWR} and Waymo \cite{Sun2020ScalabilityIP} datasets, notably achieving around 10 mAP absolute improvement over labeled-only 3D baseline on the 1\% and 2\% KITTI settings and a 10.6 AP improvement for Pedestrians in 3D on the 1\% Waymo setting.
\end{itemize}
\label{sec:introduction}
\section{Related Work}

\noindent\textbf{Semi-Supervised Learning.} SSL methods either use consistency regularization \cite{bachman2014learning,rasmus2015semi,sajjadi2016regularization,laine2016temporal,tarvainen2017mean} or pseudo-labeling \cite{lee2013pseudo,berthelot2019mixmatch,berthelot2019remixmatch,sohn2020fixmatch,Zhang2021FlexMatchBS}. The former forces noised predictions on unlabeled images to be consistent. The seminal work \cite{bachman2014learning} enforces consistency over dropout, Temporal Ensembling \cite{laine2016temporal} stores exponential moving averages (EMA) of past predictions, and Mean Teacher \cite{tarvainen2017mean} enforces consistency between ``student'' and ``teacher'' models, the latter an EMA of the former. 

Pseudo-labeling methods explicitly generate labels on unlabeled data and train on them in lieu of ground truth. MixMatch \cite{berthelot2019mixmatch} ensembles over augmentations, ReMixMatch \cite{berthelot2019remixmatch} uses weak augmentations for labeling and strong augmentations for training, and FixMatch \cite{sohn2020fixmatch} uses a confidence threshold to generate labels. Our method builds on intuitions from Mean Teacher \cite{tarvainen2017mean} and asymmetric augmentations \cite{berthelot2019mixmatch,sohn2020fixmatch} to ensure the teacher model can correctly supervise the student by maintaining an advantage over the student.

\noindent\textbf{SSL for Object Detection.} 2D detection models \cite{Ren2015FasterRT,Lin2017FeaturePN,Liu2016SSDSS,Redmon2016YouOL,Cai2018CascadeRD,Tian2019FCOSFC} consist of a feature extraction backbone \cite{He2016DeepRL}, a region proposal network \cite{Ren2015FasterRT,Liu2016SSDSS}, and optionally, a second-stage proposal refinement module \cite{Ren2015FasterRT,Cai2018CascadeRD}. 3D object detection methods \cite{Zhou2018VoxelNetEL,Graham20183DSS,Yan2018SECONDSE,Shi2020PVRCNNPF,Liang2020RangeRCNNTF} follow a similar structure, instead using voxel \cite{Graham20183DSS,Yan2018SECONDSE,Choy20194DSC,Shi2020FromPT} or point \cite{Qi2017PointNetDH,Shi2019PointRCNN3O,Qi2019DeepHV,Yang20203DSSDP3} representations instead of 2D modules. Our proposed DetMatch is agnostic to the single-modality detectors used. 

Some 2D SSL object detection methods \cite{Jeong2019ConsistencybasedSL,Tang2021HumbleTT} enforce consistency over augmentations, STAC \cite{Sohn2020ASS} generates pseudo-labels offline, and Instant-Teaching \cite{Zhou2021InstantTeachingAE} experiments with Mosaic \cite{Bochkovskiy2020YOLOv4OS} and MixUp \cite{Zhang2018mixupBE}. A line of work \cite{Xu2021EndtoEndSO,Li2021RethinkingPL} improves confidence thresholding, and other methods use EMA for predictions \cite{Yang2021InteractiveSW} and teacher models \cite{Tang2021HumbleTT,Liu2021UnbiasedTF}. Similarly, for 3D SSL, SESS \cite{Zhao2020SESSSS} performs consistency regularization over asymmetric augmentations and 3DIoUMatch \cite{Wang20213DIoUMatchLI} thresholds on predicted IoU. Unlike 2D detection, more 3D methods use offline labeling \cite{Caine2021PseudolabelingFS,Qi2021Offboard3O,Wang2021Semisupervised3O}, with some \cite{Qi2021Offboard3O,Wang2021Semisupervised3O} using extensive augmentation ensembling and multiple timesteps to refine single-frame 3D detections. Improvements in multi-frame fusion are perpendicular to our work, as our DetMatch generates cleaner per-frame pseudo-labels that can be used in place of single-modality detections for downstream multi-timestep aggregation and refinement. Unlike these single-modality SSL methods, our pipeline jointly leverages the unique characteristics of RGB images and point clouds to improve SSL for each modality.

\noindent\textbf{2D-3D Multi-Modal Learning.} Many works have explored 2D-3D fusion for detection and segmentation. Some methods \cite{Lahoud20172DDriven3O,Qi2018FrustumPF,Wang2019FrustumCS} constrain the 3D search space through 2D detection, while others fuse 2D and 3D features \cite{Sindagi2019MVXNetMV,Huang2020EPNetEP,Yoo20203DCVFGJ,Wang2021VPFNetVF,Zhao2021LIFSegLA} or predictions \cite{Qi2020ImVoteNetB3,Xie2020PIRCNNAE,Vora2020PointPaintingSF,Yin2021MultimodalVP,Park2021MultiModalityTC}. Some works have explored cross modal distillation \cite{Chong2022MonoDistillLS}, contrastive pretraining \cite{Liu2021LearningF2,Liu2020P4ContrastCL,Liu20213Dto2DDF}, or directly transferring 2D model into 3D \cite{xu2021image2point}. Most relevant to our work is xMUDA \cite{Jaritz2020xMUDACU}, which proposes a cross-modality loss for semantic segmentation domain adaptation. Their 3D model is supervised by 2D segmentation results and vice versa. However, unlike pixels and points on which segmentation is done, detections in 2D and 3D do not have a directly calculable bijective mapping, making cross-modal supervision in object detection a less constrained problem. Further, training box regression requires extra consideration. We address these difficulties in our framework and leverage the EMA teacher-student with asymmetric augmentation to stabilize training.

\label{sec:related}
\section{Method}
\subsection{Problem Definition}
In semi-supervised object detection, we have a small set of labeled data $\{(\textbf{x}_i^l, \textbf{y}_i^l)\}_{i=1}^{N_l}$ and a larger set of unlabeled data $\{\textbf{x}_i^u\}_{i=1}^{N_u}$, where $N_l$ and $N_u$ are the number of labeled and unlabeled frames, respectively. We typically have $N_u >> N_l$. We omit the scripts on $\textbf{x}_i^l$ when they are clear from context. In autonomous driving \cite{Geiger2012AreWR,Sun2020ScalabilityIP} and indoor scene understanding \cite{Song2015SUNRA,Xiao2013SUN3DAD,Janoch2011AC3,Silberman2012IndoorSA,Dai2017ScanNetR3}, a single input sample is a multi-modal tuple $\textbf{x} = (\textbf{x}_{2D}, \textbf{x}_{3D})$. $\textbf{x}_{2D}$ is a 2D RGB image
and $\textbf{x}_{3D}$ is a 3D point cloud.
Similarly, each ground truth annotation is a tuple of 2D and 3D labels, which in turn are each a set of boxes and classification labels:
\[\textbf{y} = \left(\textbf{y}_{2D} = \left\{(\textbf{b}_{2D}, \textbf{c}_{2D})^{(j)}\right\}, \textbf{y}_{3D} = \left\{(\textbf{b}_{3D}, \textbf{c}_{3D})^{(j)}\right\}\right)\]
$\textbf{b}_{2D} \in \mathbb{R}^4$ is a 2D box, $\textbf{b}_{3D} \in \mathbb{R}^7$ is a 3D box, and $\textbf{c} \in \{0, 1\}^C$ is a one-hot label indicating one of $C$ classes. To reduce the labeling burden for training, we generate $\textbf{y}_{2D}$ from $\textbf{y}_{3D}$ by projecting $\textbf{b}_{3D}$ to 2D to get $\textbf{b}_{2D}$ using camera parameters. Thus, our pipeline requires no 2D labels for the target dataset. 
\subsection{Teacher-Student Framework}
We use a student model $\textbf{S}$ and a teacher model $\textbf{T}$ of the same architecture. At a high level, the teacher $\textbf{T}$ generates pseudo-labels on the unlabeled data that the student $\textbf{S}$ trains on. For the teacher to correctly and stably supervise the student, the teacher must maintain an advantage over the student in terms of the performance. We accomplish this by iteratively updating and improving the teacher model through training via exponential moving average (EMA) accumulation:
\[\theta_{\textbf{T}} \leftarrow \alpha\theta_{\textbf{T}} + (1 - \alpha)\theta_{\textbf{S}} \tag{1}\]
where $\alpha$ is the EMA momentum, and the $\theta$ are the model parameters.
Unlike methods that pseudo-label offline \cite{Sohn2020ASS,Caine2021PseudolabelingFS,Qi2021Offboard3O}, our student and its EMA teacher allow for continuous improvement of pseudo-labels throughout training.
\subsection{Single-Modality Semi-Supervised Learning}\label{sec:3.3}
\noindent\textbf{Overview.} 
In this section, we outline a straightforward teacher-student, single-modality SSL approach based on the state-of-the-art 2D SSL method Unbiased Teacher \cite{Liu2021UnbiasedTF}. We find that with a well-tuned confidence threshold, this simple baseline compares favorably against more complicated approaches in 3D such as 3DIoUMatch \cite{Wang20213DIoUMatchLI}. We omit modality indicators $2D$ and $3D$ for this section, because this SSL baseline is applicable to any detection model.

\noindent\textbf{Pre-training on Labeled Data.}
For the teacher to guide the student, the teacher must be able to predict reasonable bounding boxes from the start. So, we first pre-train the student on the labeled data. Let $\textbf{T}(\textbf{x}) = \hat{\textbf{y}}_{\textbf{T}} = \left\{(\hat{\textbf{b}}_\textbf{T}, \hat{\textbf{c}}_\textbf{T})^{(j)}\right\}$ and $\textbf{S}(\textbf{x}) = \hat{\textbf{y}}_{\textbf{S}} = \left\{(\hat{\textbf{b}}_\textbf{S}, \hat{\textbf{c}}_\textbf{S})^{(j)}\right\}$ denote the predictions of the teacher and the student models respectively, with each consisting of a set of bounding boxes and semantic classification probabilities. The loss on the labeled samples is:
\[ \mathcal{L}^{l} = \mathcal{L}_{loc}\left(\hat{\textbf{y}}_{\textbf{S}}^{l}, \left\{{\textbf{b}^l}^{(j)}\right\}\right) + \mathcal{L}_{cls}\left(\hat{\textbf{y}}_{\textbf{S}}^{l}, \left\{{\textbf{c}^l}^{(j)}\right\}\right) \tag{2}\label{equation:lab_loss}\]
where $\mathcal{L}_{loc}$ and $\mathcal{L}_{loc}$ represent the localization and classification losses, respectively. After the student is pre-trained to convergence, the teacher is initialized with the student weights before the SSL training begins.

\noindent\textbf{Semi-Supervised Training.}
To retain representations learned from the labeled data, we train using an equal number of labeled and unlabeled samples per batch: 
\[\mathcal{L} = \mathcal{L}^l + \lambda \mathcal{L}^u \tag{3}\label{equation:loss}\]
where $\mathcal{L}^l$ is as defined in Equation \ref{equation:lab_loss}, $\mathcal{L}^u$ is the loss on unlabeled samples and $\lambda$ is a weighting hyperparameter. To train on unlabeled data, we get box predictions from the teacher and only keep the ones with maximum classification confidence above a threshold $\tau$ as pseudo-labels. We can write the teacher's generated pseudo-labels on the unlabeled data as: 
\begin{equation}
    \begin{split}
        \hat{\textbf{y}}_{\textbf{T}}^{(>\tau)} &=
        \left\{(\hat{\textbf{b}}_\textbf{T}, \hat{\textbf{c}}_\textbf{T})^{(j)}\right\}^{(>\tau)} \\
        &= \left\{\left(\hat{\textbf{b}}_{\textbf{T}}^{(j)}, \hat{\textbf{c}}_{\textbf{T}}^{(j)}\right) \Big|~ \text{max}(\hat{\textbf{c}}_{\textbf{T}}^{(j)}) > \tau \right\} 
    \end{split}\tag{4}
\end{equation}
giving us the unlabeled loss:
\begin{equation}
    \begin{split}
        \mathcal{L}^u = &\mathcal{L}_{loc}\left(\hat{\textbf{y}}_{\textbf{S}}^{u}, \left\{\textbf{b}_{\textbf{T}}^{(j)}\right\}^{(>\tau)}\right) \\
        &+ \mathcal{L}_{cls}\left(\hat{\textbf{y}}_{\textbf{S}}^{u}, \left\{\text{argmax}(\textbf{c}_{\textbf{T}}^{(j)})\right\}^{(>\tau)}\right)
    \end{split}\tag{5}\label{equation:unlab_loss}
\end{equation}
After SSL training, we take the teacher as our final model for more stability.

\noindent\textbf{Asymmetric Data Augmentation.}
\begin{figure*}[t]
  \centering
  \includegraphics[width=\linewidth]{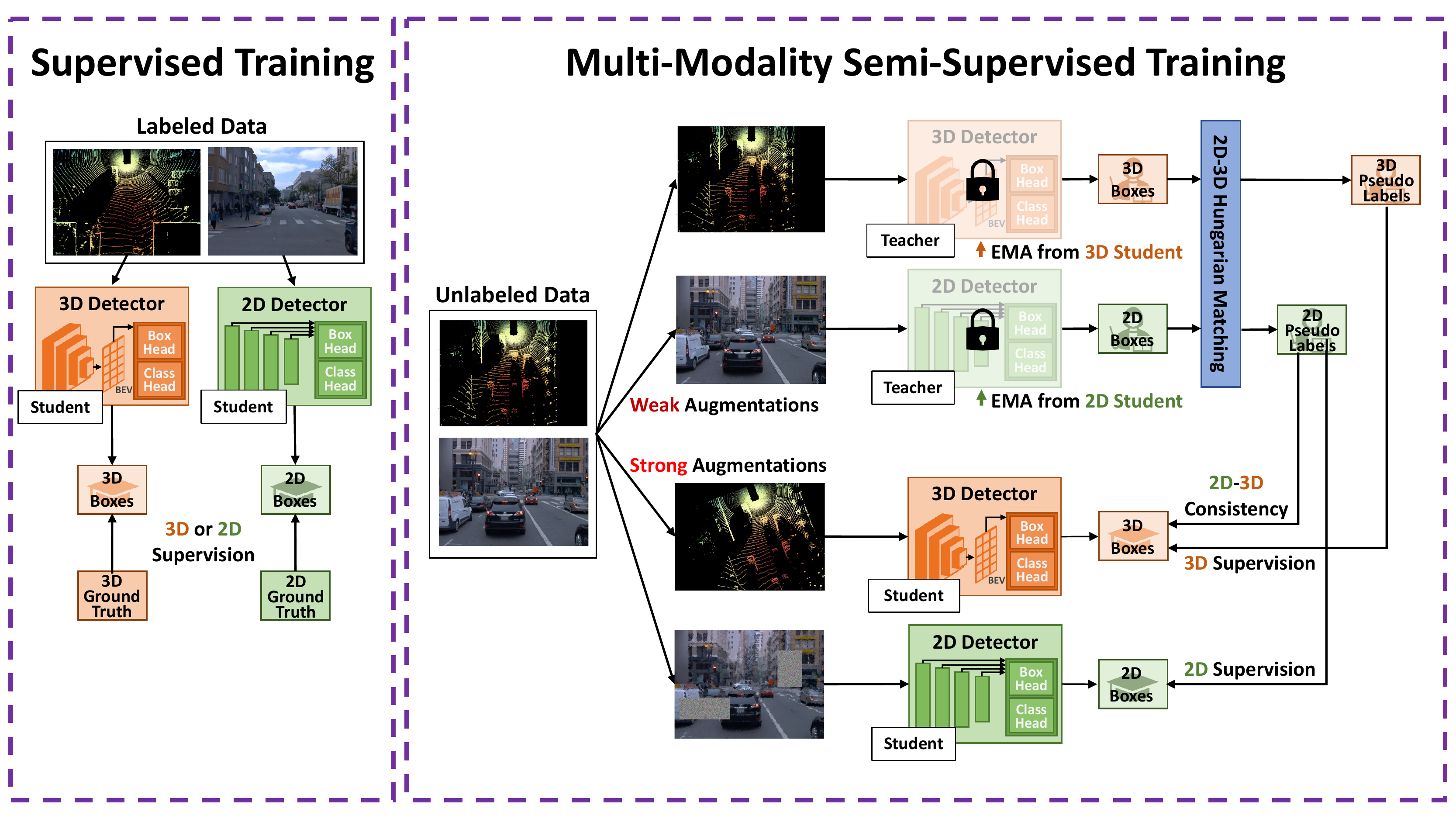}
  \caption{The proposed DetMatch. We have a teacher and student for each modality and match 2D and 3D teacher predictions to supervise the students. The 2D teacher also directly supervises the 3D student through 2D-3D Consistency}
  \label{fig:model}
  \vspace{-1em}
\end{figure*}
Although EMA makes the teacher more stable than the student, EMA alone does not give the teacher a large enough advantage in performance over the student. To further decouple their predictions, we adopt asymmetric data augmentation on the inputs of the teacher and the student. We use weak augmentation $\mathcal{A}_{weak}(\textbf{x})$ for the teacher and strong augmentation $\mathcal{A}_{strong}(\textbf{x})$ for the student.
When the teacher's pseudo-labels supervise the student, we invert geometric augmentations in $\mathcal{A}_{weak}(\textbf{x})$ and apply those in $\mathcal{A}_{strong}(\textbf{x})$ to align the space of the teacher and student boxes. 
We find that this single-modality SSL framework outperforms 3DIoUMatch on autonomous driving datasets, so we adopt it as our strong baseline for comparison.

\subsection{Multi-Modality Semi-Supervised Learning}
\noindent\textbf{Overview.} 
Although this single-modality SSL framework improves over labeled-only training, it has several disadvantages. Firstly, it does not leverage the paired 2D and 3D inputs, leading to sub-optimal single-modality results. Secondly, classification confidence is a poor measure of box localization performance as noted by prior work \cite{Jiang2018AcquisitionOL,Song2020RevisitingTS}. Finally, we find that single-modality self-training is prone to error propagation, leading to decreased performance in some cases. 

To address these problems, we present our multi-modal semi-supervised framework shown in Figure \ref{fig:model}. DetMatch jointly maintains a teacher and a student for each modality and matches 2D and 3D teacher predictions to generate a cleaner set of pseudo-labels. Furthermore, to leverage the unique advantages of dense, colorful 2D RGB images, we propose a 2D-3D consistency module that forces 3D student predictions to be similar to 2D teacher boxes. Our multi-modal framework also performs pre-training and keeps labeled losses $\mathcal{L}^l_{2D}, \mathcal{L}^l_{3D}$ during SSL training for each modality as in Section \ref{sec:3.3}. As the pseudo-label generation changes, our unlabeled losses $\mathcal{L}^u_{2D}, \mathcal{L}^u_{3D}$ are different from Equation \ref{equation:unlab_loss}. We also introduce an additional $\mathcal{L}_{consistency}$ loss. The overall loss for our DetMatch is:
\[
\mathcal{L} = (\mathcal{L}^l_{2D} + \mathcal{L}^l_{3D}) + (\mathcal{L}^u_{2D} + \mathcal{L}^u_{3D}) + \mathcal{L}_{consistency}\tag{6}
\]

\noindent\textbf{2D-3D Hungarian Matching \& Supervision.} 
\begin{figure*}[t]
  \centering
  \resizebox{0.9\textwidth}{!}{\includegraphics[width=\linewidth]{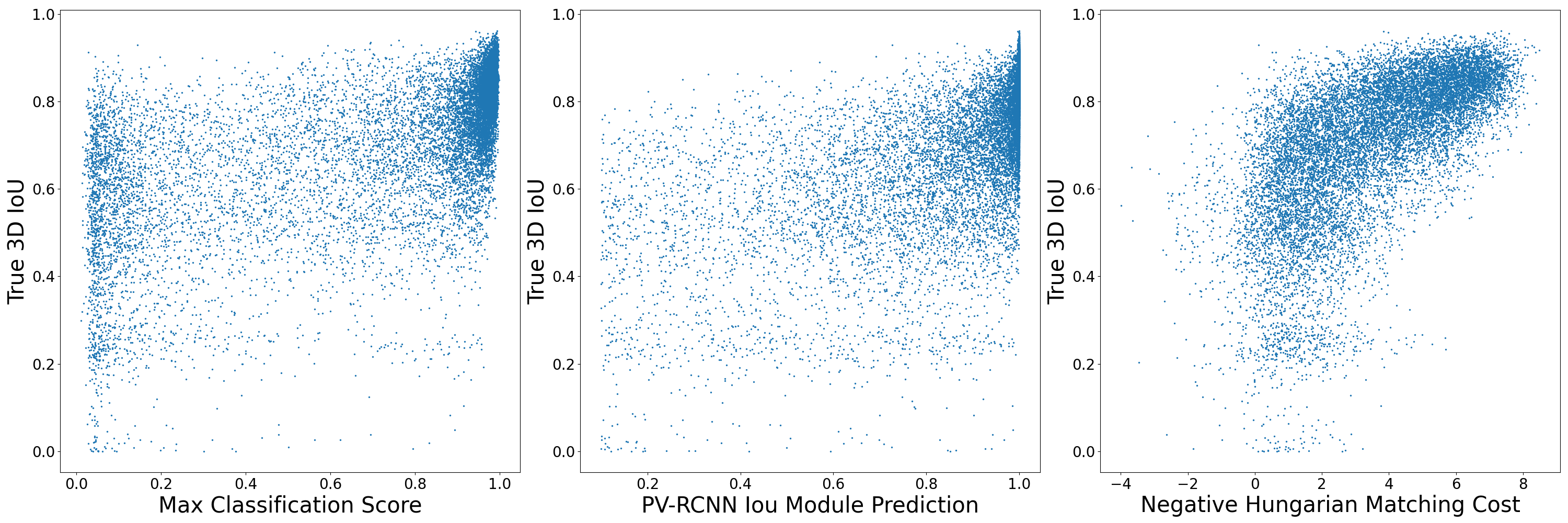}}
  \caption{Comparison between boxes' true 3D ground-truth IoU and various methods of assessing box quality on KITTI 1\% unlabeled data}
  \label{fig:iou}
  \vspace{-1em}
\end{figure*}
\begin{figure*}[t]
  \centering
  \resizebox{0.8\textwidth}{!}{\includegraphics[width=\linewidth]{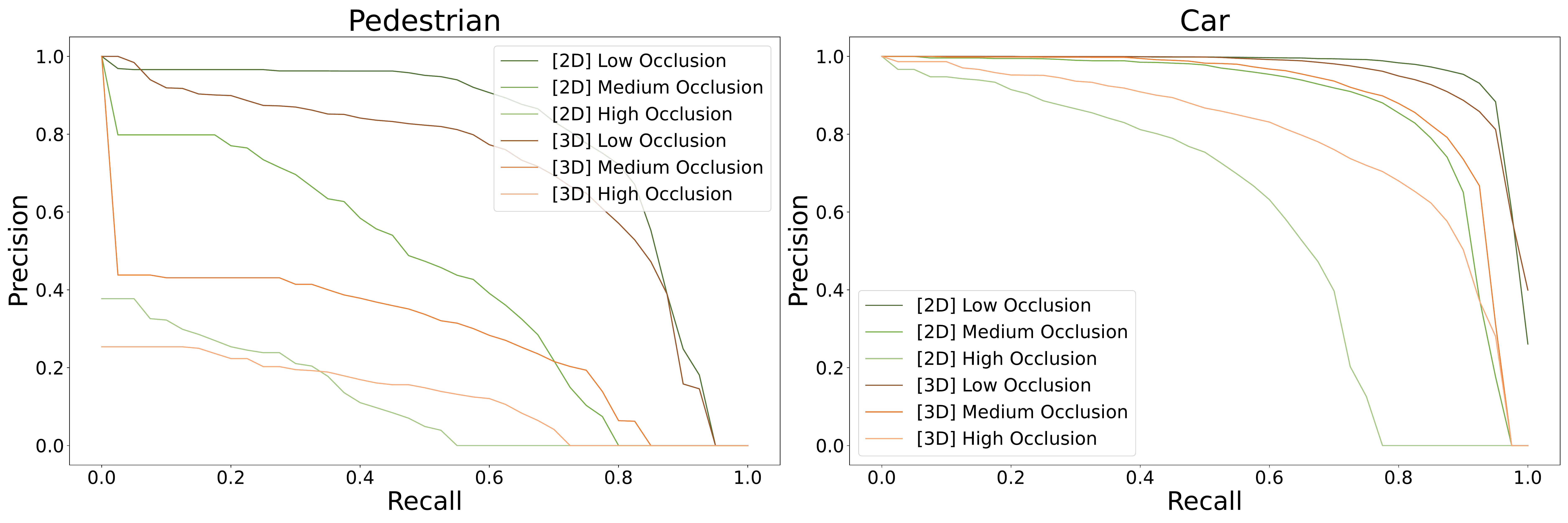}}
  \caption{2D and 3D model performance at various occlusion levels}
  \label{fig:occlusion}
  \vspace{-1em}
\end{figure*}
\begin{figure*}[t]
  \centering
  \resizebox{0.88\textwidth}{!}{\includegraphics[width=\linewidth]{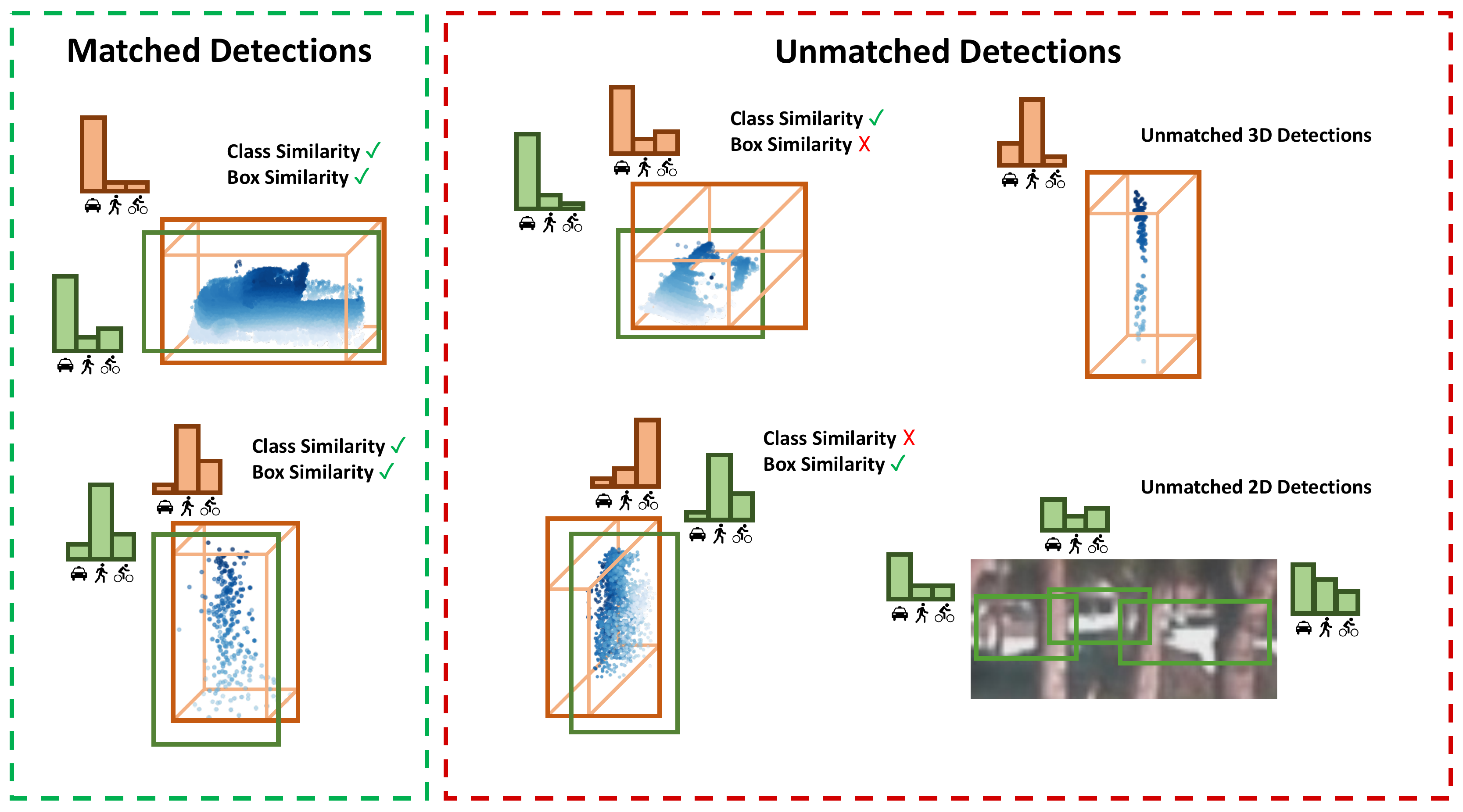}}
  \caption{Illustration of the 2D-3D Hungarian matching algorithm }
  \label{fig:hungarian}
  \vspace{-1em}
\end{figure*}
A drawback of the pipeline in Section \ref{sec:3.3} is its use of classification confidence to determine pseudo-labels. We visualize this problem in the left plot of Figure \ref{fig:iou}, which shows that many 3D boxes with a low max score are highly overlapped with a ground truth box. Moreover, although scoring modules directly supervised by true IoU \cite{Shi2020PVRCNNPF,Wang20213DIoUMatchLI} are better than max classification score as shown in the middle plot, this IoU prediction module is unable to differentiate among high IoU values 0.6 - 0.9 as evidenced by the vertical cluster on the right side. As such, pseudo-labels generated using these single-modality measures of box quality prediction remain noisy.

We first examine the pros and cons of the 2D and 3D modalities. We plot in Figure \ref{fig:occlusion} the P/R curves of 2D and 3D detections for Pedestrian and Car classes on the KITTI validation dataset, with a separate curve for ground truth objects labeled as low, medium, and high occlusion. We find that at the same occlusion level, 2D better detects and localizes Pedestrians when compared to 3D. Due to the sparsity of point clouds and their lack of color information, Pedestrians are often confused with poles and trees of similar shape in 3D. However, such ambiguous objects are clearly identifiable in the dense 2D RGB image.

On the other hand, 2D detection struggles with highly overlapping objects due to its lack of depth information - when viewed in the 3D point cloud, such overlapping objects are clearly separated.
This trend is especially clear when viewing the P/R curves for the Car class. Although 2D outperforms 3D for objects of low occlusion, we see a clear reversal for highly occluded objects. These observations clearly demonstrate that 2D and 3D modalities are complementary at the detection level - a relationship we propose to leverage for SSL by choosing as pseudo-labels detections with a corresponding match in the other modality.

More specifically, as shown in Figure \ref{fig:hungarian}, we compute an optimal bipartite matching between 2D and 3D teacher predictions using the Hungarian Algorithm \cite{Kuhn1955TheHM} and consider pairs with a matching cost below a threshold $\tau_{hung}$ ``matched''. The algorithm for matched pairs generation can be written as:
\begin{equation}
    \begin{split}
    \hspace{-3em}\left\{\left((\hat{\textbf{b}}_{\textbf{T}}, \hat{\textbf{c}}_{\textbf{T}})^{2D}, (\hat{\textbf{b}}_{\textbf{T}}, \hat{\textbf{c}}_{\textbf{T}})^{3D}\right)^{(j)}\right\}^{(<\tau_{hung})} \\
    = Hungarian_{2D\text{-}3D}^{\tau_{hung}}(\hat{\textbf{y}}_{\textbf{T}}^{2D}, \hat{\textbf{y}}_{\textbf{T}}^{3D})
    \end{split}
     \tag{7}\label{equation:hung} 
\end{equation}
We omit notation for the matching algorithm and thresholding for brevity. Inspired by recent works \cite{Carion2020EndtoEndOD,Sun2021SparseRE} on detection using learnable queries, our matching cost between a pair of 2D and 3D box predictions has three components:
\begin{equation}
    \begin{split}
        \mathcal{L}_{match} &\left((\hat{\textbf{b}}_{\textbf{T}}, \hat{\textbf{c}}_{\textbf{T}})^{2D}, (\hat{\textbf{b}}_{\textbf{T}}, \hat{\textbf{c}}_{\textbf{T}})^{3D}\right) \\
        &=\lambda_{L1} \mathcal{L}_{L1} + \lambda_{iou} \mathcal{L}_{iou} + \lambda_{d\text{-}focal} \mathcal{L}_{d\text{-}focal}
    \end{split}
     \tag{8}\label{equation:hung_match} 
\end{equation}
Note that unlike classification score, a \textit{lower} cost indicates a stronger match. 

$\mathcal{L}_{L1}$ and $\mathcal{L}_{iou}$ are box consistency costs between the projected 3D box and the 2D box. To get the former, we use camera parameters to project the 8 corners of the 3D box to the image and compute a tightly fitted 2D box. $\mathcal{L}_{L1}$ calculates $l_1$ loss between the 2D box parameters and $\mathcal{L}_{iou}$ calculates generalized IoU loss \cite{Rezatofighi2019GeneralizedIO} between the 2D boxes. These costs force paired 2D and 3D pseudo-labels to refer to the same object and be in agreement regarding the precise localization that object. Unlike single-modality box localization confidence methods that suffer from modality-specific drawbacks and self-confidence bias, our multi-modal box consistency cost gives us a natural way to assess box quality.

$\mathcal{L}_{d\text{-}focal}$ calculates class prediction consistency between the 2D and 3D predictions. We formulate a double-sided version of FocalLoss \cite{Lin2020FocalLF}:
\begin{equation}
    \begin{split}
        \mathcal{L}_{d\text{-}focal} = &FocalLoss\left(\hat{\textbf{c}}_{\textbf{T}}^{2D}, \text{argmax}(\hat{\textbf{c}}_{\textbf{T}}^{3D})\right) \\
        &+ FocalLoss\left(\hat{\textbf{c}}_{\textbf{T}}^{3D}, \text{argmax}(\hat{\textbf{c}}_{\textbf{T}}^{2D})\right)
    \end{split}
    \tag{9}\label{equation:doublefocal}
\end{equation}
Note that this double-sided FocalLoss allows for a smooth trade-off between 2D and 3D confidence. A low-confidence 3D box \textit{can still be chosen as a pseudo-label} if its matched 2D box has high confidence. Intuitively, high-confidence predictions of one modality can ``promote'' low-confidence predictions of the other modality, a dynamic selection not possible with simple confidence thresholding. Further, although this formulation of $\mathcal{L}_{focal}$ does prefer higher-confidence boxes, its motivation is different from that of confidence thresholding - $\mathcal{L}_{focal}$ considers \textit{consistency} between classification predictions in 2D and 3D. If both modalities agree on the semantic class of a region, they will have a lower matching cost. 

Our proposed 2D-3D matching cost is a remarkably more accurate measure of box localization quality as shown in the rightmost plot of Figure \ref{fig:iou}. We then use the matched and thresholded pairs of 2D and 3D teacher boxes as pseudo-labels to supervise the 2D and 3D students on the unlabeled data:
\begin{align*}
    \mathcal{L}^u_{modal} =& \mathcal{L}_{loc}\left(\hat{\textbf{y}}_{\textbf{S}}^{modal}, \left\{(\textbf{b}_{\textbf{T}}^{modal})^{(j)}\right\}^{(<\tau_{hung})}\right) + \\
    & \mathcal{L}_{cls}\left(\hat{\textbf{y}}_{\textbf{S}}^{modal}, \left\{\text{argmax}\left((\textbf{c}_{\textbf{T}}^{modal})^{(j)}\right)\right\}^{(<\tau_{hung})}\right) \\
    & \text{for } modal \in \{2D, 3D\} \tag{10}\label{equation:hung_unlab_loss}
\end{align*}

\noindent\textbf{2D-3D Consistency.} 
\begin{figure*}[t]
  \centering
  \includegraphics[width=\linewidth]{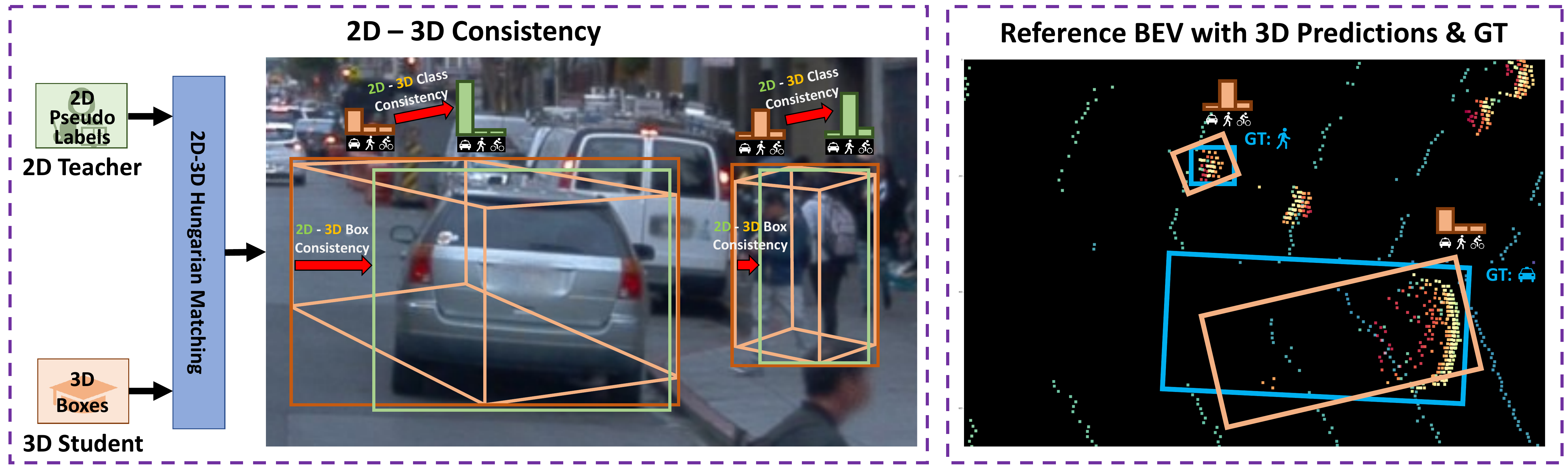}
  \caption{The box and class consistency between the 2D teacher and the 3D student}
  \label{fig:consistency}
  \vspace{-1em}
\end{figure*}
Through our 2D-3D Hungarian Matching, we generated a cleaner set of pseudo-labels to supervise each student. However, although we have leveraged the advantages 3D can provide 2D, we have not fully exploited the benefits 3D can get from 2D. 
We have fulfilled the former because although a core advantage of 3D is detection of highly occluded or visually unclear boxes, we need to differentiate these beneficial 3D teacher boxes from the false positives 3D detection is especially prone to. So, it is necessary to first match 3D boxes with 2D teacher boxes to filter noisy boxes while retaining the beneficial boxes.

On the other hand, the dense, semantically rich format of 2D RGB images make class confusion less likely and instead enables better localization of non-heavily occluded objects as shown in Figure \ref{fig:occlusion}. So, compared to 3D, high-confidence 2D boxes are less likely to be false positives, allowing them to directly rectify incorrect 3D boxes. 
However, in our previous pipeline, 2D teacher boxes can only supervise 3D indirectly through 3D teacher boxes that are potentially worse than 2D in terms of classification and localization. We propose to directly match 2D teacher boxes and 3D student boxes and enforce box and class consistency between them as shown in Figure \ref{fig:consistency}. Applying Hungarian Matching and thresholding as in Equation \ref{equation:hung}:
\begin{equation}
    \begin{split}
        \hspace{-3em}\left\{\left((\hat{\textbf{b}}_{\textbf{T}}, \hat{\textbf{c}}_{\textbf{T}})^{2D}, (\hat{\textbf{b}}_{\textbf{S}}, \hat{\textbf{c}}_{\textbf{S}})^{3D}\right)^{(j)}\right\}^{(<\tau_{hung})} \\
        = Hungarian_{2D\text{-}3D}^{\tau_{hung}}(\hat{\textbf{y}}_{\textbf{T}}^{2D}, \hat{\textbf{y}}_{\textbf{S}}^{3D}) 
    \end{split}
    \tag{11}\label{equation:hung_consistency} 
\end{equation}
Then, the 2D-3D consistency loss between matched 2D and 3D pairs is:
\[
\mathcal{L}_{consistency} = \lambda_{L1} \mathcal{L}_{L1} + \lambda_{iou} \mathcal{L}_{iou} + \lambda_{focal} \mathcal{L}_{focal}\tag{12}\label{equation:consistency}
\]
Losses $\mathcal{L}_{L1}$ and $\mathcal{L}_{iou}$ are identical to the box consistency costs in Equation \ref{equation:hung_match}. $\mathcal{L}_{focal}$ is the regular FocalLoss with the 3D student probabilities supervised by the semantic class of the 2D teacher box. This final 2D-3D consistency loss allows us to fully utilize the strengths of 2D to improve our multi-modal framework.

\label{sec:method}
\section{Experiments}
\subsection{Datasets and Evaluation Metrics}
\noindent\textbf{KITTI.}
Following 
3DIoUMatch, we evaluate on 1\% and 2\% labeled data sampled over 3712 training frames, and
we also validate our method by sampling 20\% of driving \textit{sequences}. 
We average results over three sampled splits for each \% setting, using the released 3DIoUMatch splits for 1\% and 2\%. We report for both 2D and 3D the moderate mAP for the Car, Pedestrian, and Cyclist classes.

\noindent\textbf{Waymo Open Dataset.}
We also evaluate on the large-scale Waymo dataset, which has 158361 training frames. 
Each frame has 360 degree LiDAR and 5 RGB cameras, with the cameras only capturing 240 degrees. This limitation, coupled with the complex and diverse urban setting, makes multi-modal training especially difficult on Waymo. We validate our framework on the 1\% labeled data setting, sampling 1\% of the 798 sequences, which results in around 1.4k frames. Due to the sheer scale of the Waymo dataset and the observation that even this 1\% split has four times the cars and eight times the pedestrians as the full KITTI dataset, we validate on a single Waymo split. We report mAP and mAPH at both LEVEL 1 and LEVEL 2 difficulties for Car and Pedestrian.

\subsection{Implementation Details}
We use PV-RCNN \cite{Shi2020PVRCNNPF} for 3D detection and Faster-RCNN \cite{Ren2015FasterRT} with FPN \cite{Lin2017FeaturePN} and ResNet50 \cite{He2016DeepRL} backbone for 2D detection. To reduce labeling costs specifically for autonomous driving, we follow multi-modality methods \cite{Qi2018FrustumPF,Sindagi2019MVXNetMV,Vora2020PointPaintingSF} and initialize the 2D detector with weights pre-trained on COCO \cite{Lin2014MicrosoftCC}. This is a reasonable setting because labeling costs associated with annotating autonomous driving frames in 3D for specific applications do not preclude the existence of publicly available 2D detection datasets in another domain. Further, we relax this assumption in Table \ref{table:no_coco} by evaluating DetMatch without COCO pre-training and find that our method still dramatically improves over SSL baselines.

We set $\tau_{3D} = 0.3$, $\tau_{2D} = 0.7$, and $\tau_{hung} = -1.5$, and use the same $\tau_{hung}$ threshold for both applications of Hungarian Matching. For KITTI, we train for 5k iterations with a batch size of 24; for Waymo, we train for 12k iterations with a batch size of 12. Additional details can be found in the supplementary.

\subsection{Results on KITTI}
\begin{table*}[t]

\begin{center}
\resizebox{\textwidth}{!}{
    \begin{tabular}{c||cccc|cccc||cccc}
        \multirow{2}{*}{Method} & \multicolumn{4}{c|}{1\%} & \multicolumn{4}{c||}{2\%} & \multicolumn{4}{c}{20\%}\\
        & mAP & Car & Ped & Cyc & mAP & Car & Ped & Cyc & mAP & Car & Ped & Cyc \\
        \hline
        \begin{tabular}{@{}c@{}}{\footnotesize Labeled-Only} \\ {\footnotesize (3DIoUMatch Reported)} \end{tabular} & 43.5 & 73.5 & 28.7 & 28.4 & 54.3 & 76.6 & 40.8 & 45.5 & - & - & - & -\\
        \hline
        3DIoUMatch & 48.0 & 76.0 & 31.7 & 36.4 & 61.0 & 78.7 & 48.2 & 56.2 & - & - & - & - \\
        Improvement & +4.5 & +2.5 & +3.0 & +8.0 & +6.7 & \textbf{+2.1} & +7.4 & +10.7 & - & - & - & - \\
        \hline
        \hline
        \begin{tabular}{@{}c@{}}{\footnotesize Labeled-Only} \\ {\footnotesize (Reproduced by Us)} \end{tabular} & 45.9 & 73.8 & 30.4 & 33.4 & 55.8 & 76.1 & 44.9 & 46.4 & 61.3 & 77.9 & 47.1 & 58.9 \\
        \hline
        Confidence Thresholding & 54.4 & 75.9 & 42.7 & 44.6 & 63.3 & 76.5 & 50.0 & 63.4 & 68.1 & 77.8 & 58.0 & 68.6 \\
        Improvement & +8.5 & +2.1 & +12.3 & \textbf{+11.2} & +7.5 & +0.4 & +5.1 & +17.0 & +6.8 & -0.1 & \textbf{+10.9} & +9.7 \\
        \hline
        Ours & 59.0 & 77.5 & 57.3 & 42.3 & 65.6 & 78.2 & 54.1 & 64.7 & 68.7 & 78.7 & 57.6 & 69.6 \\
        Improvement & \textbf{+13.1} & \textbf{+3.7} & \textbf{+26.9} & +8.9 & \textbf{+9.8} & \textbf{+2.1} & \textbf{+9.2} & \textbf{+18.3} & \textbf{+7.4} & \textbf{+0.8} & +10.5 & \textbf{+10.7} 
    \end{tabular}
}
\caption{3D detection performance comparison on KITTI. Training on the labeled samples to convergence, we observe slightly better labeled-only performance than 3DIouMatch. Improvement is increase from labeled-only results.}
\label{table:kitti_3d}
\end{center}
\vspace{-1em}
\end{table*}

\begin{table*}[t]
\begin{center}
\resizebox{\textwidth}{!}{
    \begin{tabular}{c||cccc|cccc||cccc}
        \multirow{2}{*}{Method} & \multicolumn{4}{c|}{1\%} & \multicolumn{4}{c||}{2\%} & \multicolumn{4}{c}{20\%}\\
        & mAP & Car & Ped & Cyc & mAP & Car & Ped & Cyc & mAP & Car & Ped & Cyc \\
        \hline
        Labeled-Only & 65.3 & 86.6 & 68.6 & 40.8 & 68.9 & 87.4 & 70.7 & 48.3 & 63.9 & 87.5 & 64.5 & 39.8 \\
        \hline
        Confidence Thresholding & 60.4 & 86.1 & 69.2 & 25.8 & 65.5 & 87.6 & 71.5 & 37.2 & 66.2 & 88.8 & 70.0 & 39.7 \\
        Improvement & -4.9 & -0.5 & +0.6 & -15.0 & -3.4 & +0.2 & +0.8 & -11.1 & +2.3 & +1.3 & +5.5 & -0.1 \\
        \hline
        Ours & 71.4 & 88.8 & 73.9 & 51.7 & 74.5 & 89.0 & 74.6 & 59.9 & 72.8 & 89.1 & 71.6 & 57.7 \\
        Improvement & \textbf{+6.1} & \textbf{+2.2} & \textbf{+5.3} & \textbf{+10.9} & \textbf{+5.6} & \textbf{+1.6} & \textbf{+3.9} & \textbf{+11.6} & \textbf{+8.9} & \textbf{+1.6} & \textbf{+7.1} & \textbf{+17.9}
    \end{tabular}
}
\caption{2D detection performance comparison on KITTI. Note that although we train with projected 3D boxes, we evaluate with annotated 2D boxes.}
\label{table:kitti_2d}
\end{center}
\vspace{-1em}
\end{table*}
We evaluate our model on 2D and 3D object detection on KITTI, comparing with 3DIoUMatch and our SSL baseline, which is equivalent to Unbiased Teacher \cite{Liu2021UnbiasedTF} in 2D. The results are shown in Tables \ref{table:kitti_3d} and \ref{table:kitti_2d}. First, we find that with a well-tuned 3D confidence threshold, our 3D-only confidence thresholding baseline is able to outperform 3DIoUMatch in both mAP absolute performance and improvement. However, we note that for the Car class, 3DIoUMatch outperforms the 3D SSL baseline which struggles to improve performance over labeled-only training in 2\% and 20\% settings. This is because Car is the most common class and is already well-trained just from the labeled data, making further improvements difficult. Our proposed DetMatch, leveraging both 2D and 3D detections, consistently outperforms all methods. Notably, we find that in the 1\% setting, we observe a remarkable \textbf{26.9\%} boost in AP, far outperforming 3DIoUMatch, which achieves a 3\% improvement, and our 3D SSL baseline, which achieves a 12.3\% improvement. This gap can be attributed to the ambiguity of pedestrians in 3D and the relative clarity of this class when viewed in the RGB image.

For 2D detection, we see that the Unbiased Teacher baseline suffers from a drop in performance through SSL training for 1\% and 2\% settings despite our hyperparameter search. We attribute this to two factors. First, SSL on autonomous driving datasets is a more difficult setting than SSL on COCO because driving datasets like KITTI have less image diversity, making it more susceptible to over-fitting. Indeed, as the amount of labeled data increases for KITTI, 2D SSL improves. We note that even the limited 1\% setting on COCO has 1171 images, each in a completely different scene. On the other hand, KITTI 1\% only has 37 images, and even the larger 20\% setting, due to its constraint of sampling driving sequences, has comparatively lower scene diversity. These factors, coupled with pre-training on COCO which strengthens the original model, make improving on the labeled-only baseline difficult. Second, single-modality training is far more susceptible to self-training error propagation. Although the asymmetric augmentation and EMA work to decouple the student from the teacher, their predictions are still highly correlated, causing the student to overfit to its own predictions, including its own errors. Our results show that the proposed DetMatch is more robust to these factors, demonstrating substantial performance gains over the labeled-only and 2D SSL baselines. Notably, we find that detection of Cyclists, a rare category, declines by 15\% mAP under Unbiased Teacher in KITTI 1\% but improves by 10.9\% mAP with DetMatch, a gap of 25.9\% mAP.

\subsection{Results on Waymo Open Dataset}
\begin{table*}[t]

\begin{center}
\resizebox{\textwidth}{!}{
    \begin{tabular}{c||cc|cc|cc|cc||cc|cc}
        \multirow{3}{*}{1\% Data} & \multicolumn{8}{c||}{3D} & \multicolumn{4}{c}{2D} \\
        \cline{2-13}
        & \multicolumn{2}{c|}{Car L1} & \multicolumn{2}{c|}{Car L2} & \multicolumn{2}{c|}{Ped L1} & \multicolumn{2}{c||}{Ped L2} & \multicolumn{2}{c|}{Car} & \multicolumn{2}{c}{Ped} \\
        & mAP & mAPH & mAP & mAPH & mAP & mAPH & mAP & mAPH & L1 & L2 & L1 & L2\\
        \hline
        Labeled-Only & 47.3 & 45.6 & 43.6 & 42.0 & 28.9 & 15.6 & 26.2 & 14.1 & 42.3 & 39.5 & 50.8 & 47.0 \\
        \hline
        Confidence Thresholding & 52.6 & 51.6 & 48.4 & 47.5 & 35.2 & 16.7 & 32.0 & 15.2 & 44.4 & 41.3 & 48.7 & 45.1 \\
        Improvement & \textbf{+5.3} & \textbf{+6.0} & \textbf{+4.8} & \textbf{+5.5} & +6.3 & +1.1 & +5.8 & +1.1 & +2.1 & +1.8 & -2.1 & -1.9\\
        \hline
        Ours & 52.2 & 51.1 & 48.1 & 47.2 & 39.5 & 18.9 & 35.8 & 17.1 & 47.8 & 44.4 & 50.6 & 46.8 \\
        Improvement & +4.9 & +5.5 & +4.5 & +5.2 & \textbf{+10.6} & \textbf{+3.3} & \textbf{+9.6} & \textbf{+3.0} & \textbf{+5.5} & \textbf{+4.9} & -0.2 & -0.2
    \end{tabular}
}
\caption{Performance comparison on the validation set of the Waymo Dataset.}
\label{table:waymo}
\end{center}
\vspace{-2em}
\end{table*}
To test the robustness of our framework, we additionally benchmark DetMatch on the difficult Waymo dataset. Because Waymo's 2D cameras have a combined FOV of 240 degrees, we use the 3D SSL pseudo-labels for the remaining 120 degrees when training DetMatch. We keep hyperparameters of DetMatch, which were tuned on KITTI, the same for Waymo and find that they are generally applicable. Our 3D and 2D results are summarized in Table \ref{table:waymo}. We find that the confidence thresholding baseline is strong, consistently demonstrating improvements of 5\% or 6\% on the mAP metric for 3D. For 2D, we see a smaller improvement and even observe the performance on pedestrian drop by two points. We attribute this to the same factors that caused a drop in KITTI - although Waymo dataset is larger, its 1\% labeled data diversity less than that of COCO. 

DetMatch slightly drops in performance for Cars in 3D compared to the SSL baseline. However, it improves on the SSL baseline by a substantial 4.3 mAP for Pedestrian L1. Further, DetMatch achieves a large boost of 3.4 mAP for Car L1 in 2D over single-modality SSL, and although it does not boost performance for 2D Pedestrian, DetMatch stymies the decline from Unbiased Teacher. 

Overall, compared to the labeled-only and SSL baselines, our method significantly boosts performance for Pedestrian on 3D and Car on 2D while largely maintaining other settings' performance. We attribute the large Pedestrian 3D improvement to DetMatch's effective use of RGB images' advantage in identifying and localizing this class. On the other hand, the Car 2D boost stems from the 2D detector benefiting from 3D's stronger detection of Cars, which are often highly occluded in the urban streets captured in Waymo. Thus, although our DetMatch does not uniformly boost all classes, perhaps due to Faster-RCNN with ResNet50 being an older and weaker model in 2D compared to PV-RCNN in 3D, the remarkable boost regardless in Pedestrian 3D detection and Car 2D detection demonstrate that our pipeline is effective in exploiting the unique advantages of each sensor to improve detections of the other modality.

\subsection{Ablation Studies and Discussion}
In this section, we focus on quantitative results. Extensive visualizations demonstrating the pseudo labels generated by our model can be found in the supplementary.

\noindent\textbf{Threshold for DetMatch.}
\begin{table*}[t]
    \begin{minipage}[t]{.33\textwidth}
        \vspace{-6em}
        \centering
        \resizebox{\textwidth}{!}{
            \begin{tabular}{c||cccc}
                3D Eval & mAP & Car & Ped & Cyc \\
                \hline
                Labeled-Only & 45.9 & 73.8 & 30.4 & 33.4  \\
                \hline 
                $\tau_{hung} = -1$ & 54.2 & 76.1 & 49.3 & 37.2  \\
                $\tau_{hung} = -1.5$ & \textbf{57.9} & 76.7 & \textbf{55.0} & \textbf{42.0} \\
                $\tau_{hung} = -2$ & 52.4 & \textbf{76.9} & 43.7 & 36.7  \\
            \end{tabular}
        }
        \caption{3D Effect of $\tau_{hung}$}
        \label{table:hung_thresh_3d}
        \centering
        \resizebox{\textwidth}{!}{
            \begin{tabular}{c||cccc}
                2D Eval & mAP & Car & Ped & Cyc \\
                \hline
                Labeled-Only & 65.3 & 86.6 & 68.6 & 40.8  \\
                \hline 
                $\tau_{hung} = -1$ & 69.3 & 87.9 & 70.4 & 49.5 \\
                $\tau_{hung} = -1.5$ & \textbf{70.2} & 88.7 & \textbf{72.1} & \textbf{49.9} \\
                $\tau_{hung} = -2$ & 56.5 & \textbf{89.5} & 52.3 & 27.7 \\
            \end{tabular}
        }
        \caption{2D Effect of $\tau_{hung}$}
        \label{table:hung_thresh_2d}
    \end{minipage}
    \hfill
    \begin{minipage}[t]{.65\textwidth}
        \centering
        \resizebox{\textwidth}{!}{
            \begin{tabular}{c||cccc|cccc}
                \multirow{2}{*}{1\% Data} & \multicolumn{4}{c|}{3D} & \multicolumn{4}{c}{2D} \\
                \cline{2-9}
                & mAP & Car & Ped & Cyc & mAP & Car & Ped & Cyc \\
                \hline
                Labeled-Only & 45.9 & 73.8 & 30.4 & 33.4 & 65.3 & 86.6 & 68.6 & 40.8  \\
                
                \textcolor{gray} {+Confidence Thresholding} & \textcolor{gray} {54.4} & \textcolor{gray} {75.9} & \textcolor{gray} {42.7} & \textcolor{gray} {44.6} & \textcolor{gray} {60.4} & \textcolor{gray} {86.1} & \textcolor{gray} {69.2} & \textcolor{gray} {25.8} \\
                
                + {\footnotesize 2D-3D Teacher Matching} & 57.9 & 76.7 & 55.0 & 42.0 & 70.2 & 88.7 & 72.1 & 49.9 \\
                + \begin{tabular}{@{}c@{}}{\scriptsize 2D Teacher \& 3D Student} \\ {\scriptsize Box Consistency} \end{tabular} & 59.4 & 77.4 & 56.5 & 44.4 & 69.8 & 88.5 & 71.9 & 49.0 \\
                + \begin{tabular}{@{}c@{}}{\scriptsize 2D Teacher \& 3D Student}\\ {\scriptsize Class Consistency} \end{tabular} & 59.0 & 77.5 & 57.3 & 42.3 & 71.4 & 88.8 & 73.9 & 51.7 \\
                
                \textcolor{gray} {+} \begin{tabular}{@{}c@{}}{\scriptsize \textcolor{gray} {2D Teacher \& 3D Student}} \\ {\scriptsize \textcolor{gray} {MSE instead of Focal}} \end{tabular} & \textcolor{gray} {58.2} & \textcolor{gray} {77.6} & \textcolor{gray} {57.7} & \textcolor{gray} {39.3} & \textcolor{gray} {68.1} & \textcolor{gray} {88.6} & \textcolor{gray} {72.0} & \textcolor{gray} {50.8} \\
            \end{tabular}
        }
        \caption{Ablation of DetMatch Modules}
        \label{table:ablation}
    \end{minipage}
\end{table*}
Results for KITTI 1\% at various $\tau_{hung}$ on DetMatch with just the 2D-3D Teacher Matching pseudo-labeling module are shown in Tables \ref{table:hung_thresh_3d} and \ref{table:hung_thresh_2d}. Ablations on single-modality thresholds $\tau_{3D}$ and $\tau_{2D}$ are in the supplementary. We find that Car prefers a more stringent (lower) cost threshold. Further, we observe that 2D and 3D mAP both peak at the \textit{same} $\tau_{hung} = -1.5$, which shows that improvements in one modality strongly benefit the other.

\noindent\textbf{Ablation of Multi-Modal Components.}
Next, we study the effect of each module of DetMatch in Table \ref{table:ablation}. Components not part of our final model are in gray. We focus on the Car and Pedestrian classes for this fine-grained comparison as Cyclist results vary by up to 3 AP even on 100\% labeled data runs. Replacing the single-modality thresholding with our 2D-3D teacher matched pseudo-labels results in a large improvement. This shows us that pseudo-labeling with objects consistently detected in both modalities better supervises the student. 

Enforcing box consistency between the 2D teacher and 3D student improves substantially improves the 3D performance with a small 0.2 point drop in Car and Pedestrian 2D performance. We attribute this boost to the 3D student now generating boxes that better fit objects in the dense 2D image. FocalLoss class consistency boosts 3D and 2D Pedestrian performance by 0.8 and 2 points, respectively. This is in-line with our observations that Pedestrian is difficult to detect in 3D - by rectifying class prediction of under-confident or incorrect 3D detections using 2D, the 3D model improves. Further, the 2D performance improves because 2D pseudo-labels are tied with 3D teacher predictions. By training the 3D model to generate more accurate 3D Pedestrian detections, the 2D model is better supervised as well. This improvement demonstrates the mutually beneficial relationship between improvements in the 2D and 3D models. 

We try replacing FocalLoss in class consistency with MSE following Mean Teacher \cite{tarvainen2017mean}. That this decreases performance gives us more insight into the purpose of class consistency. MSE encourages logit matching \cite{tarvainen2017mean,Kim2021ComparingKD}, which is closely related to knowledge distillation \cite{Hinton2015DistillingTK}, where, by imitating class similarities predicted by a teacher, the student learns the underlying function of the teacher. In our setting, the teacher and student are of different modalities and consume data of very different representations, inhibiting such mimicking. As such, what our consistency module does is directly interpretable - it rectifies 3D student box and class predictions using the 2D teacher outputs.

\noindent\textbf{Without COCO Pre-training.}
\begin{table*}[t]

\begin{center}
\resizebox{.8\textwidth}{!}{
    \begin{tabular}{cc||cccc|cccc}
        \multicolumn{2}{c||}{\multirow{2}{*}{1\% Data}} & \multicolumn{4}{c|}{3D} & \multicolumn{4}{c}{2D} \\
        \cline{3-10}
        & & mAP & Car & Ped & Cyc & mAP & Car & Ped & Cyc \\
        \hline
        \multirow{2}{*}{\begin{tabular}{@{}c@{}}{\scriptsize w/ COCO }\\ {\scriptsize Pre-Training} \end{tabular}} & \multicolumn{1}{|c||}{Labeled-Only} & 45.9 & 73.8 & 30.4 & 33.4 & 65.3 & 86.6 & 68.6 & 40.8  \\
        & \multicolumn{1}{|c||}{Ours} & 59.0 & 77.5 & 57.3 & 42.3 & 71.4 & 88.8 & 73.9 & 51.7 \\
        \hline
        \multirow{2}{*}{\begin{tabular}{@{}c@{}}{\scriptsize w/o COCO }\\ {\scriptsize Pre-Training} \end{tabular}} & \multicolumn{1}{|c||}{Labeled-Only} & 45.9 & 73.8 & 30.4 & 33.4 & 46.2 & 77.6 & 47.1 & 13.9 \\
        & \multicolumn{1}{|c||}{Ours} & 57.1 & 77.7 & 55.3 & 38.3 & 59.1 & 85.9 & 59.0 & 30.7 
    \end{tabular}
}
\caption{Impact of COCO Pre-training}
\label{table:no_coco}
\end{center}
\vspace{-1em}
\end{table*}
We also evaluate our pipeline without COCO pre-training, as shown in Table \ref{table:no_coco}. We find that although COCO pre-training is important for 2D performance, we still achieve strong 3D performance without it, notably maintaining a substantial 24.9\% AP improvement for Pedestrian. This shows that DetMatch does not need COCO, instead benefiting more from the multi-modal interaction. Further, improvements from using COCO shows that our framework is a unique and effective way of transferring benefits from 2D labels, which are easier to annotate than 3D labels, to the 3D detection task.

\label{sec:experiments}
\section{Conclusion}
In this work, we proposed DetMatch, a flexible multi-modal SSL framework for object detection that obtains state-of-the-art performance on various limited labeled data settings on KITTI and Waymo. We demonstrate that pseudo-labels generated by matching 2D and 3D detections allow each modality to benefit from the other's advantages and improvements. Further, by enforcing consistency between 3D student and 2D teacher boxes, we leverage the unique advantages that the dense RGB image gives the 2D detector in detecting ambiguous objects. As our pipeline achieves improved performance on 3D detection by using a COCO pre-trained 2D detector, our method also shows potential in leveraging cheaper or publicly available 2D annotations to lower 3D data requirements.

\label{sec:conclusion}
{\small
\bibliographystyle{ieee_fullname}
\bibliography{egbib}
}

\clearpage
\appendix
\section*{Supplementary}

\section{Overview}
In this supplementary, we provide additional training details, quantitative results, and qualitative visualizations. These sections are organized as follows:
\begin{itemize}
    \item Section \ref{sec:train_details} provides additional details about our implementation and training setup.
    \item Section \ref{sec:SSL_thresholds} contains the ablation study for the thresholds of the single-modality 2D and 3D SSL pipelines.
    \item Section \ref{sec:second} presents additional quantitative results with DetMatch using SECOND instead of PV-RCNN, demonstrating the adaptability of our method.
    \item Section \ref{sec:visualizations} contains additional visualizations of DetMatch on KITTI and Waymo.
\end{itemize}
\section{Additional Implementation and Training Details}
\label{sec:train_details}
\begin{table*}[t]

\begin{center}
\resizebox{.8\textwidth}{!}{
    \begin{tabular}{c||cccc|cccc}
        \multirow{2}{*}{1\% Data} & \multicolumn{4}{c|}{3D} & \multicolumn{4}{c}{2D} \\
        \cline{2-9}
        & mAP & Car & Ped & Cyc & mAP & Car & Ped & Cyc \\
        \hline 
        Labeled-Only & 38.3 & 65.4 & 22.6 & 26.9 & 65.3 & 86.6 & 68.6 & 40.8 \\
        \hline 
        Confidence Thresholding & 38.8 & 70.1 & 26.7 & 19.7 & 60.4 & 86.1 & 69.2 & 25.8 \\
        Improvement & +0.5 & +4.7 & +4.1 & -7.2 & -4.9 & -0.5 & +0.6 & -15.0 \\
        \hline 
        Ours & 49.4 & 74.9 & 41.9 & 31.5 & 68.5 & 88.7 & 70.3 & 46.5 \\
        Improvement & \textbf{+11.1} & \textbf{+9.5} & \textbf{+19.3} & \textbf{+4.6} & \textbf{+3.2} & \textbf{+2.1} & \textbf{+1.7} & \textbf{+5.7} 
    \end{tabular}
}
\caption{KITTI Results for DetMatch + SECOND}
\label{table:second_kitti}
\end{center}
\vspace{-1em}
\end{table*}
\begin{table*}[t]

\begin{center}
\resizebox{\textwidth}{!}{
    \begin{tabular}{c||cc|cc|cc|cc||cc|cc}
        \multirow{3}{*}{1\% Data} & \multicolumn{8}{c||}{3D} & \multicolumn{4}{c}{2D} \\
        \cline{2-13}
        & \multicolumn{2}{c|}{Car L1} & \multicolumn{2}{c|}{Car L2} & \multicolumn{2}{c|}{Ped L1} & \multicolumn{2}{c||}{Ped L2} & \multicolumn{2}{c|}{Car} & \multicolumn{2}{c}{Ped} \\
        & mAP & mAPH & mAP & mAPH & mAP & mAPH & mAP & mAPH & L1 & L2 & L1 & L2\\
        \hline
        Labeled-Only & 35.6 & 34.4 & 32.6 & 31.5 & 19.7 & 10.4 & 17.8 & 9.4 & 42.3 & 39.5 & 50.8 & 47.0 \\
        \hline
        Confidence Thresholding & 42.7 & 41.8 & 40.1 & 39.3 & 27.7 & 13.3 & 25.1 & 12.1 & 44.4 & 41.3 & 48.7 & 45.1 \\
        Improvement & +7.1 & +7.4 & +7.5 & +7.8 & +8.0 & +2.9 & +7.3 & +2.7 & +2.1 & +1.8 & -2.1 & -1.9\\
        \hline
        Ours & 45.2 & 44.1 & 41.5 & 40.6 & 35.7 & 16.9 & 32.3 & 15.3 & 48.1 & 44.8 & 51.1 & 47.1 \\
        Improvement & \textbf{+9.6} & \textbf{+9.7} & \textbf{+8.9} & \textbf{+9.1} & \textbf{+16.0} & \textbf{+6.5} & \textbf{+14.5} & \textbf{+5.9} & \textbf{+5.8} & \textbf{+5.3} & \textbf{+0.3} & \textbf{+0.1}
    \end{tabular}
}
\caption{Waymo Results for DetMatch + SECOND}
\label{table:second_waymo}
\end{center}
\vspace{-1em}
\end{table*}
For all settings, including single-modality pipelines and DetMatch, we pre-train on the labeled data, then initialize both the teacher and student model with these weights to begin SSL training. Similar to prior work \cite{Zhao2020SESSSS,Wang20213DIoUMatchLI}, we ramp-up EMA momentum from 0.99 to 0.999. We follow Unbiased Teacher \cite{Liu2021UnbiasedTF} in only supervising classification for the 2D detector in both single-modality SSL and DetMatch. Further, we adopt their setting of maintaining a constant learning rate through training and taking the teacher as the final model - we find that this yields more stable and reproducible performance. Moreover, we replace the classification loss in the 2D detector with Focal Loss \cite{Lin2020FocalLF}, which has been found \cite{Liu2021UnbiasedTF} to yield more class-balanced pseudo-labels in SSL. The 3D detectors PV-RCNN \cite{Shi2020PVRCNNPF} and SECOND \cite{Yan2018SECONDSE} use Focal Loss by default, so they are left unchanged. For fair comparison, we use the same 2D weak-strong augmentations as Unbiased Teacher and the same 3D weak-strong augmentations as 3DIoUMatch. All experiments were conducted on three NVIDIA A6000 GPUs.

\subsection{KITTI Training Details}
\noindent\textbf{Pre-training.} Owing to the smaller number of samples, we pre-train the 2D detector for 240 epochs and the 3D detector for 800 epochs on the 1\%, 2\% settings and for 120 and 400 epochs on the 20\% setting. An epoch is defined as one pass over the limited labeled data, and we find that these long cycles allow for full convergence. The batch size is 24 for both 2D and 3D detectors. The 2D detector is trained using SGD with starting learning rate 0.03 decayed 10x twice mimicking the standard ``1x'' COCO training cycle. PV-RCNN is trained using AdamW \cite{Loshchilov2019DecoupledWD} with the one-cycle scheduling strategy and a max learning rate of 0.015. SECOND is trained with the same settings and PV-RCNN but with a max learning rate of 0.0045

\noindent\textbf{SSL Training.} Each SSL batch consists of 12 labeled and 12 unlabeled samples, and we use the starting or max learning rate of pre-training as the constant learning rate when training 2D or 3D SSL. Based on observations in other multi-modality works \cite{Zhang2020MultiModalityCA}, we use a separate optimizers for 2D and 3D in DetMatch, maintaining SGD for 2D and AdamW for 3D.

\subsection{Waymo Training Details}
\noindent\textbf{Pre-training.} We use the Waymo v1.0 released data. We pre-train the 2D detector for a standard ``1x'' COCO training cycle on the 1\% setting and train the 3D detector for 48 epochs. Due to the much higher resolution of Waymo's 2D images, we do not do multiscale training, instead keeping the original 1920x1280 resolution. Additionally, due to GPU memory limitations, a single 2D sample in Waymo consists of the front view and a side view, the latter sampled from one of the four side images. We do this because the front view has far more objects on average than the other views. Finally, the detectors are trained with half the batch size and learning rate as they were in KITTI. 

\noindent\textbf{SSL Training.} Each SSL batch consists of 6 labeled and 6 unlabeled samples. We find that for Waymo, the raw LiDAR intensity value wildly varies from 0 to tens of thousands, causing instability in the early layers. As such, we freeze the first block for 3D detectors during SSL training. For DetMatch, the 2D teacher predicts boxes on all five 2D views for 2D-3D teacher Hungarian Matching, but only two 2D views are used to train the 2D student due to memory limitations. Since the five 2D images have a combined FOV of 240 degrees, we simply use confidence thresholding on the 3D teacher to generate pseudo-labels on the remaining 120 degrees. Despite not being able to apply DetMatch on the full 3D scene, we find that our pipeline improves over the 3D SSL baseline.
\section{Thresholds for 2D and 3D Single-Modality SSL}
\label{sec:SSL_thresholds}
\begin{table}[t]
    \begin{minipage}[t]{\linewidth}
        \centering
        \resizebox{\textwidth}{!}{
            \begin{tabular}{c||cccc}
                3D Eval & mAP & Car & Ped & Cyc \\
                \hline
                Labeled-Only & 45.9 & 73.8 & 30.4 & 33.4 \\
                \hline 
                $\tau_{3d} = 0.2$ & 48.6 & 75.2 & 33.4 & 37.1 \\
                $\tau_{3d} = 0.3$ & \textbf{54.4} & 75.9 & \textbf{42.7} & \textbf{44.6} \\
                $\tau_{3d} = 0.4$ & 50.6 & \textbf{76.4} & 35.0 & 40.3 \\
                $\tau_{3d} = 0.5$ & 45.7 & 72.7 & 31.4 & 42.7 \\
            \end{tabular}
        }
        \caption{Impact of $\tau_{3D}$}
        \label{table:thresh_3d}
    \end{minipage}
    \hfill
    \begin{minipage}[t]{\linewidth}
        \centering
        \resizebox{\textwidth}{!}{
            \begin{tabular}{c||cccc}
                2D Eval & mAP & Car & Ped & Cyc \\
                \hline
                Labeled-Only & 65.3 & 86.6 & 68.6 & 40.8 \\
                \hline 
                $\tau_{3d} = 0.6$ & 55.7 & 84.1 & 67.6 & 15.4 \\
                $\tau_{3d} = 0.7$ & \textbf{60.4} & 86.1 & \textbf{69.2} & \textbf{25.8} \\
                $\tau_{3d} = 0.8$ & 57.5 & \textbf{88.0} & 60.4 & 24.3  \\
            \end{tabular}
        }
        \caption{Impact of $\tau_{2D}$}
        \label{table:thresh_2d}
    \end{minipage}
\end{table}

We extensively search for the best confidence thresholds $\tau_{3D}, \tau_{2D}$ for our 3D and 2D SSL baselines. The results are shown in Tables \ref{table:thresh_3d} and \ref{table:thresh_2d}. We observe that for both modalities, although the best mAP is acheived at $\tau_{3D}=0.3$ and $\tau_{2D}=0.7$, Car detection peaks at a slightly higher threshold. 3DIoUMatch adopts different thresholds for Car, but to avoid introducing additional hyperparameters, we use a single threshold for both SSL baselines and DetMatch. It is worth noting, however, that even if we had used a class-specific threshold for Car for the single-modality SSL baselines, our DetMatch still achieves better Car performance.

For 2D-only SSL, we exhaustively searched confidence thresholds, training schedules, and weighting parameters but were unable to find a setting that yields improved performance on all classes for KITTI 1\%. As mentioned in the main paper, this can be attributed to the more difficult and limited-data setting of SSL on autonomous driving datasets as well as the single-modality self-training error propagation. Indeed, we find that on KITTI 20\% results shown in the main paper, 2D SSL is able to improve performance, demonstrating that more labeled data is required to improve on the already-strong 2D labeled-only baseline.
\section{DetMatch with SECOND}
\label{sec:second}
To demonstrate the adaptability of DetMatch, we replace the two-stage PV-RCNN 3D detector with a representative one-stage 3D detector SECOND. The KITTI results are shown in Table \ref{table:second_kitti} and the Waymo results are in Table \ref{table:second_waymo}.

We find that on KITTI, although confidence thresholding is able to substantially improve 3D Car and Pedestrian results, it reduces performance for Cyclist. Since Cyclist is a rare category, with only a dozen samples in the KITTI 1\% setting, we find that SECOND, a weaker but substantially faster 3D detector than PV-RCNN, is not able to sufficiently learn this class to generate accurate pseudo-labels for self-training. Our DetMatch addresses this problem, more accurately identifying high quality pseudo-labels by considering consistency between 2D and 3D detections. DetMatch substantially improves performance in all metrics. Further, unlike other approaches \cite{Wang20213DIoUMatchLI,Li2021RethinkingPL}, DetMatch does not require additional modules to estimate box quality, being readily adaptable to various detectors.

We see a similar trend on the Waymo dataset as shown in Table \ref{table:second_waymo}. Our DetMatch improves on both the labeled-only model and the strong single-modality SSL baselines, notably improving 3D Pedestrian performance by 16.0 mAP and 2D Car performance by 5.7 mAP. Interestingly, unlike DetMatch with PV-RCNN, we find that DetMatch with SECOND improves significantly over the 3D-only SSL baseline for the Car class. We attribute this to SECOND being a weaker detector than PV-RCNN and thus being able to derive more benefits from joint training with the 2D detector. Overall, we find that DetMatch is adaptable, able to easily work with various detectors, and that its single hyperparameter $\tau_{hung}$ is robust under various settings. This is due $\tau_{hung}$ thresholding on a \textit{consistency} cost between 2D and 3D detections, which is fundamentally different from simple confidence thresholding. Predicted class confidence is a single model's evaluation of its own predictions, subject to self-bias and error propagation through training. On the other hand, consistency cost can be considered one model ``consulting'' another to evaluate its predictions. As this cost depends on \textit{agreement} between semantic class predictions and box parameters of these two models, it is a measurement of box quality more decoupled from any single model. Drawing from these benefits, our DetMatch demonstrates improvement over labeled-only and single-modality SSL methods.
\section{Qualitative Results}
\label{sec:visualizations}
\begin{figure*}[t]
  \centering
  \includegraphics[width=\linewidth]{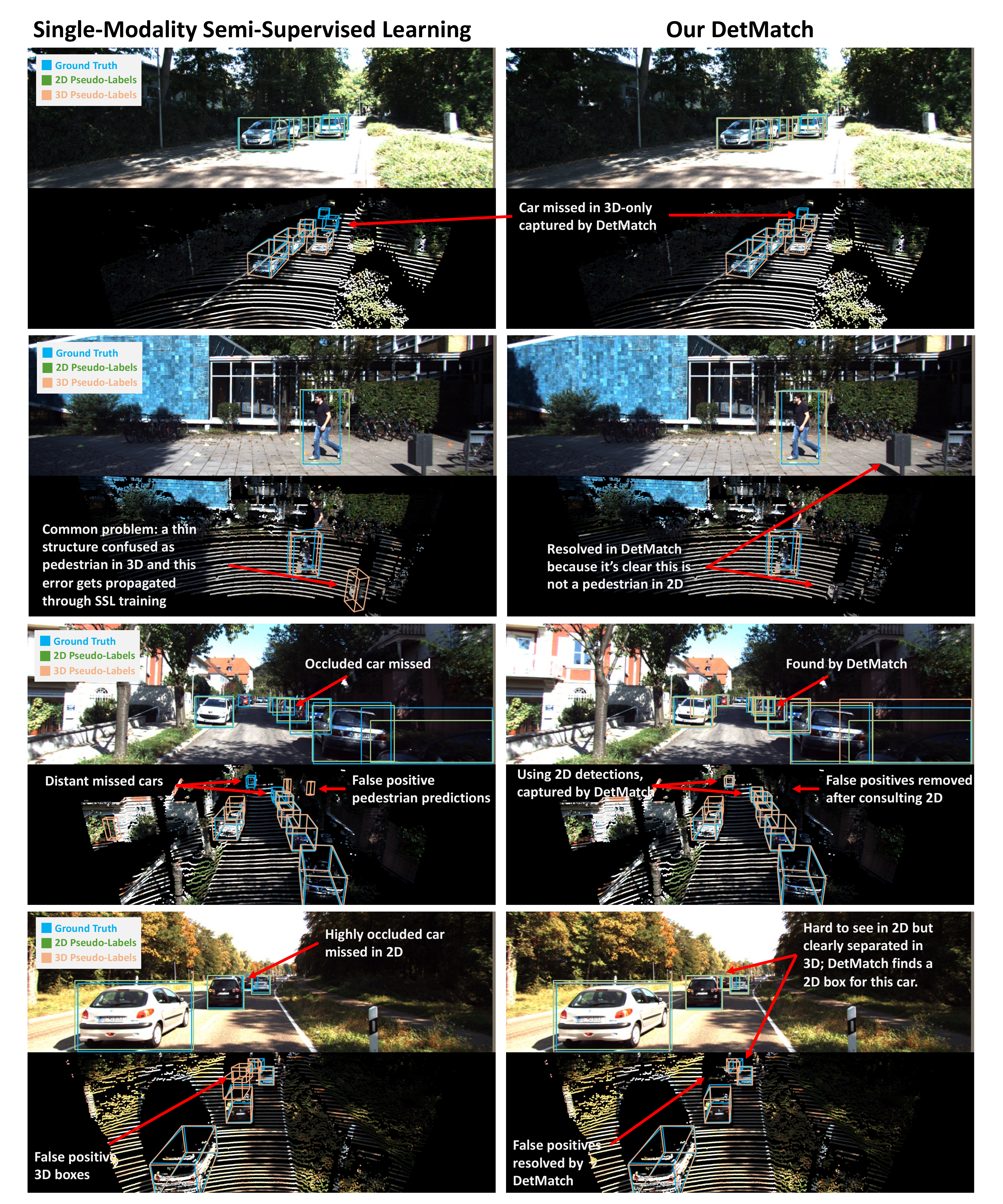}
  \caption{Visualizations of DetMatch on KITTI}
  \label{fig:kitti_vis}
  \vspace{-1em}
\end{figure*}
\begin{figure*}[t]
  \centering
  \includegraphics[width=.85\linewidth]{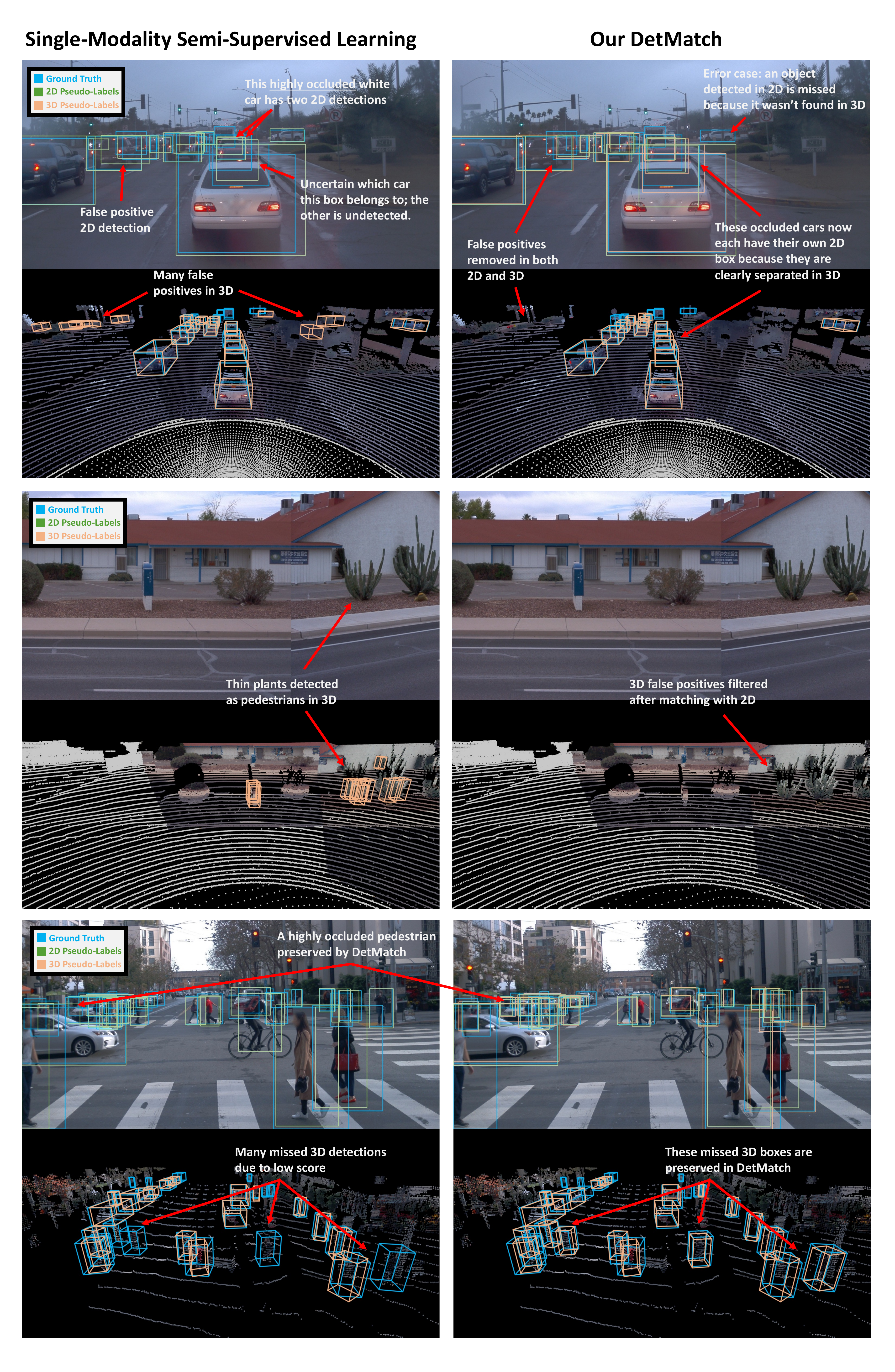}
  \vspace{-1em}
  \caption{Visualizations of DetMatch on the Waymo Dataset}
  \label{fig:waymo_vis}
  \vspace{-1em}
\end{figure*}

We present qualitative results of DetMatch for KITTI and Waymo in Figures \ref{fig:kitti_vis} and \ref{fig:waymo_vis}, respectively. Note that although point cloud is colored for visualization, 3D-only detectors take as input a color-less point cloud so that they can be applied to LiDAR-only setups as well. 

We find that our method is effective in utilizing the advantages of 2D detections to preserve correct 3D detections and vice versa. 2D detections are especially useful in removing false positive 3D detections and ``promoting'' accurate but low confidence pedestrian 3D detections that are filtered away in 3D-only SSL. On the other hand, we observe that 3D detections are better at objects highly occluded in 2D because these objects are clearly separated in 3D. This property allows DetMatch to generate clear and accurate 2D boxes even for highly overlapped cars. We also notice some error cases of DetMatch. When an object correctly detected in one modality does not have a corresponding prediction in the other modality, the correct detection is not preserved as a pseudo-label. Although this causes our method to miss some objects, DetMatch generates far fewer false positives than single-modality SSL pipelines, and many works with very high thresholds \cite{sohn2020fixmatch,Zhou2021InstantTeachingAE} have observed that such a set of precise, albeit sparser, pseudo-labels is preferable to many noisy labels. We will investigate how to leverage objects correctly detected in only one modality in future work.
\end{document}